\documentclass[letterpaper, 10 pt, conference]{ieeeconf}  

\usepackage{cite}    
\makeatletter
\let\NAT@parse\undefined
\makeatother
\usepackage[colorlinks,linkcolor=blue,citecolor=blue]{hyperref}

\usepackage{amsmath,amssymb,amsthm}
\usepackage{booktabs}
\usepackage{multirow}
\usepackage{graphicx} 
\usepackage{stfloats}
\usepackage{pifont}
\usepackage{fontawesome}
\usepackage{siunitx}

\usepackage{caption}
\usepackage{lipsum}

\IEEEoverridecommandlockouts                              

\title{\LARGE \bf
Choose What to Manipulate: Revealing Data Scaling Laws in Bounding-Box Guided Policies for Semantic Manipulation
\vspace{-0.6cm}
}

\author{Yihao Wu$^{1}$, Jinming Ma$^{2}$, Junbo Tan$^{1\dagger}$, Yanzhao Yu$^{1}$,\\
Shoujie Li$^{1}$, Mingliang Zhou$^{2}$, Diyun Xiang$^{2}$, Xueqian Wang$^{1\dagger}$
\thanks{*Work done during the internship at Xiaomi Robotics Lab.}
\thanks{$^{1}$Center for Intelligent Control and Telescience, Tsinghua Shenzhen International Graduate School, Shenzhen, China. }%
\thanks{$^{2}$Beijing Xiaomi Robot Technology Co., Ltd
602, 6th Floor, Building 5, No. 15 10th Kechuang Street, Beijing Economic-Technological Development Area, Beijing, China, 100176
        }%
\thanks{$\dagger$Corresponding author: \{tjblql, wang.xq\}@sz.tsinghua.edu.cn}
}

\usepackage{etoolbox}
\newcommand{\insertfig}{
    \includegraphics[width=\linewidth]{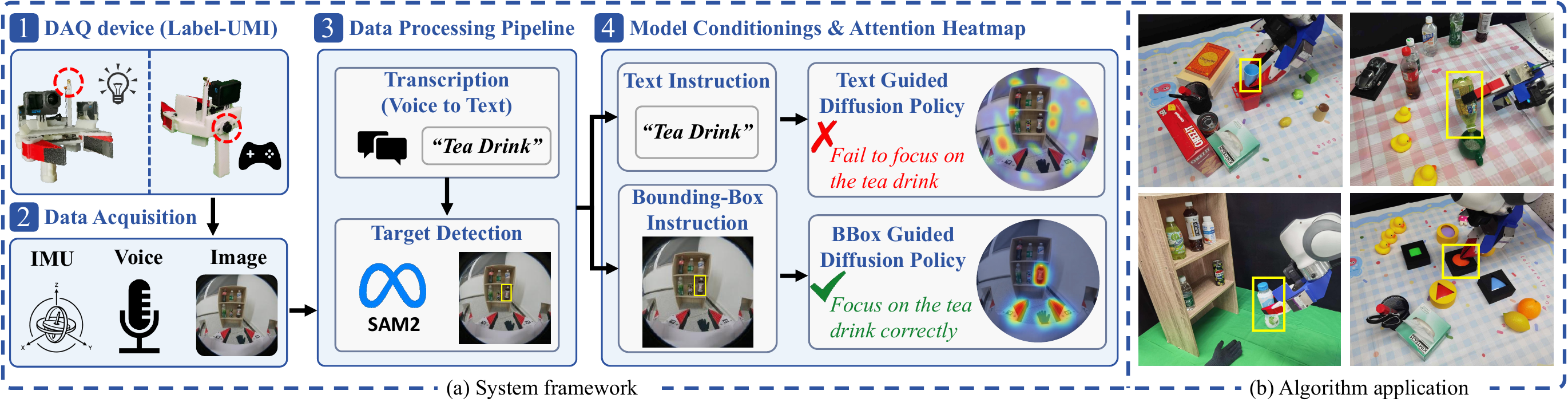}
    \captionof{figure}{Overview of the framework. (a) The Label-UMI device acquires multimodal demonstrations (IMU, voice, image); a voice command is transcribed and the target detected via SAM2, and the resulting bounding box conditions the diffusion policy, yielding sharper, more target-focused attention than text instructions. (b) Application across diverse real-world manipulation scenes.}
    \label{Overview of framework}
    \vspace{-0.6cm}
    }
    
\makeatletter
\apptocmd{\@maketitle}{\centering\insertfig}{}{}
\makeatother
\graphicspath{ {images/} }

\begin{document}

\maketitle

\thispagestyle{empty}
\pagestyle{empty}


\begin{abstract}

Diffusion-based policies generalize poorly in semantic manipulation, a key obstacle to real-world deployment, because text-only instructions cannot reliably steer the policy toward the target object in cluttered, dynamic scenes. We instead use bounding-box instructions to specify the target directly, and study how performance scales with data. To this end, we build Label-UMI, a handheld segmentation device with an automated annotation pipeline for efficiently collecting semantically labeled demonstrations, and propose a semantic-motion-decoupled framework that couples object detection with a bounding-box-guided diffusion policy; a first-frame anchoring mechanism keeps execution robust to missed detections and noisy boxes. We find that generalization follows a bounded, saturating data-scaling law with diminishing returns, validated on four real-world tasks with 6,400 demonstrations, and distill an object-diversity-first collection strategy reaching 85\% success in cluttered scenes. All data and code will be released.

\end{abstract}

\section{INTRODUCTION}

Data scaling laws \cite{kaplan2020scaling} have significantly accelerated the development of natural language processing and computer vision, as evidenced by the success of large language models (e.g., GPT \cite{achiam2023gpt}) and vision-language models (e.g., Minigpt-4\cite{zhu2023minigpt}). 
Recently, the robotics community has begun to ask whether similar data scaling behavior emerges in robotic semantic manipulation, yet research remains scarce. A fundamental obstacle is that policies relying solely on text instructions often fail at robust semantic manipulation: under identical observations, multiple manipulable objects or feasible actions may coexist, making it hard to focus on the intended target from text alone \cite{li2024virt} (Fig.\ref{Overview of framework}(a)), especially in cluttered, noisy environments. Without a reliably performing policy, studying how performance scales with data is infeasible.

To achieve reliable semantic manipulation, existing approaches can be grouped into three categories, each facing distinct scalability or applicability challenges. The first category encodes text instructions into features fused with visual inputs via FiLM or cross-attention (e.g., OCTO \cite{team2024octo}, RDT-1B \cite{liu2024rdt}); however, these methods require large-scale language-annotated demonstration datasets (e.g., millions of triplets \cite{liu2023visual}), limiting scalability. The second category leverages pre-trained LLMs for direct language interpretation (e.g., OpenVLA \cite{kim2025openvla}, OpenVLA-OFT \cite{kim2025fine}), enhancing linguistic grounding but incurring high computational costs and limited real-time performance. 
The third category employs fine-grained visual guidance such as 2D trajectory sketches or 3D keypoints to guide semantic manipulation
(e.g., Rt-Trajectory \cite{gu2023rt}, HAMSTER \cite{li2025hamster}, KITE \cite{sundaresan2023kite}); yet these often depend on specific camera perspectives, dense 3D sensing, or detailed annotations, restricting their applicability in scalable, egocentric, or minimally-instrumented settings.



In this paper, we propose a semantic-motion decoupled architecture for robotic semantic manipulation, featuring a novel collaborative reasoning mechanism between object detection models and diffusion policy models. 
As shown in Fig.~\ref{Overview of the BBox-DP}, our framework first extracts the target object from the semantic text and turns it into a bounding-box visual representation, chosen because it balances annotation efficiency with sufficient spatial guidance, unlike more intricate geometric representations.
Building on this, our framework offloads the generalization burden to an object detection module (e.g., YOLO~\cite{varghese2024yolov8} or more advanced models), and presents the detected target as a bounding-box visual instruction that the policy interprets more directly than text---consistent with biological learning \cite{piaget2013construction}, where infants acquire manipulation skills mainly through visual exploration rather than language.

In this design, the diffusion-based policy only needs to learn to follow the bounding-box visual instructions; in other words, to manipulate the object specified by a bounding box, thereby decoupling semantic grounding from motion control. 
Furthermore, we systematically investigate data scaling in semantic manipulation and find that generalization follows a bounded scaling law: it improves with the number of bounding-box objects but with clearly diminishing returns, saturating toward an upper bound rather than growing without bound. Motivated by this observation, we propose an object-diversity-first data collection strategy that substantially improves policy generalization.

Overall, our contributions can be summarized as follows:

\begin{itemize}

\item \textbf{Handheld semantic annotation device.} We design Label-UMI, a lightweight handheld segmentation device that extends the UMI system. The Label-UMI enables efficient demonstration data collection in the wild, and it provides accurate semantic annotations labels through segmentation points.

\item \textbf{Bounding-Box Guided Diffusion Policy.} We propose BBox-DP, a semantic--motion decoupled framework that couples any object detector with a diffusion policy and uses bounding boxes as visual instructions to offload generalization to the detector. A First-Frame Anchoring mechanism locks the policy to the target throughout an episode, keeping execution robust to intermittent detection failures and noisy boxes.

\item \textbf{Data scaling laws for semantic manipulation.} 
We reveal a data scaling law for bounding-box guided semantic manipulation: generalization improves with the number of bounding-box objects but with clearly diminishing returns, saturating toward an upper bound rather than growing without bound. Guided by this law, we distill an object-diversity-first data collection strategy that prioritizes diverse BBox objects over more demonstrations per object. Across four tasks with 6,400 demonstrations, the resulting policy reaches around 85\% success, validating its scalability and practicality.


\end{itemize}




\section{RELATED WORK}

\subsection{Data Collection and Diffusion Policy for Robotic Manipulation}

Robotic data collection is fundamental for training manipulation policies. Early teleoperation approaches \cite{mandlekar2018roboturk}---via VR \cite{cheng2024open} or leader--follower setups \cite{zhao2023learning}---suffer from high costs and constrained scenarios, making in-the-wild collection impractical. Recent handheld portable devices \cite{chi2024universal} enable platform-independent collection in diverse environments, yet typically lack continuous annotation of key object information (shape, position, masks) crucial for semantic manipulation.


Diffusion models are increasingly applied in robotic manipulation for their generative and generalization capabilities, e.g., visuomotor learning \cite{chi2023diffusion} and zero-shot execution \cite{black2023zero}, yet still struggle with semantic grasping and object-level generalization. We address this by integrating bounding-box guided object information into diffusion policies.

\subsection{Application of Bounding Boxes in Robot Manipulation}

Bounding boxes are widely used in vision-based grasping for localization, pose estimation, and grasp planning \cite{tan2024manibox}, providing spatial cues for downstream tasks---e.g., keypoint sampling in Im2Flow2Act \cite{xu2024flow} and a low-dimensional state in ManiBox \cite{tan2024manibox}---and, recently, high-level reasoning, serving as reasoning tokens to ground language in ECOT \cite{zawalski2024robotic} and enforcing compositional constraints in energy-based rearrangement \cite{gkanatsios2023energy}. A related line injects richer 3D representations into diffusion policies, e.g., point clouds in DP3 \cite{ze2024dp3} and category-level semantic fields in GenDP \cite{wang2024gendp}, but these need dense 3D sensing and per-category fields with high overhead. In contrast, we use 2D bounding boxes as scalable, annotation-efficient visual instructions and study the empirical scaling trend between bounding-box object diversity and policy generalization.



Most methods predict bounding boxes with YOLO variants \cite{park2020single} or custom networks \cite{zhang2019multi} such as ResNet-101 with an RPN, but labeling data is time-consuming. Some methods \cite{tan2024manibox} therefore adopt zero-shot detectors like Grounding DINO, which nonetheless perform poorly in open-ended environments and on objects described with rich adjectives. We instead propose a data acquisition and annotation method that improves efficiency while maintaining accuracy.



\section{METHOD}


In this section, we first introduce the hardware design and data processing pipeline for batch data collection. We then present an improved diffusion policy leveraging bounding box representations. Finally, we provide a formal formulation of the data scaling laws and describe our rigorous evaluation protocol.

\addtocounter{figure}{-1}
\begin{figure*}[ht]
    \centering
    \includegraphics[width=0.8\linewidth]{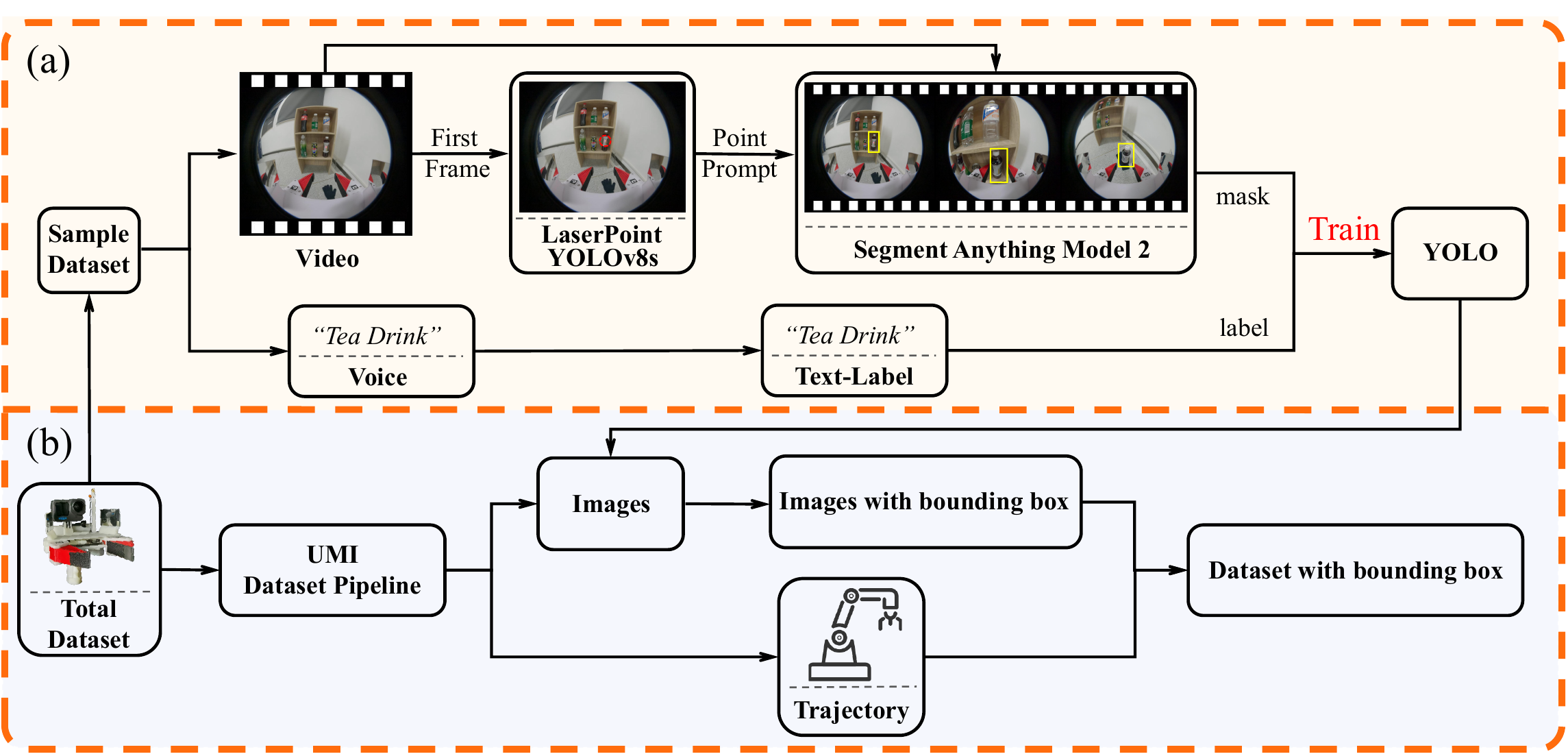}
    \caption{Overview of the data acquisition pipeline. (a) YOLO model acquisition: from a randomly sampled subset, the laser prompt and SAM2 produce per-frame bounding boxes that serve as labels to train a YOLO object detector. (b) Full-dataset annotation: the UMI pipeline extracts trajectory and image data, and the trained YOLO detector automatically annotates all frames with bounding boxes.
}
    \label{Overview of Data Acquisiton pipeline.}
    \vspace{-0.65cm}
\end{figure*}

\begin{figure}[t]
    \centering
    \includegraphics[width=\linewidth]{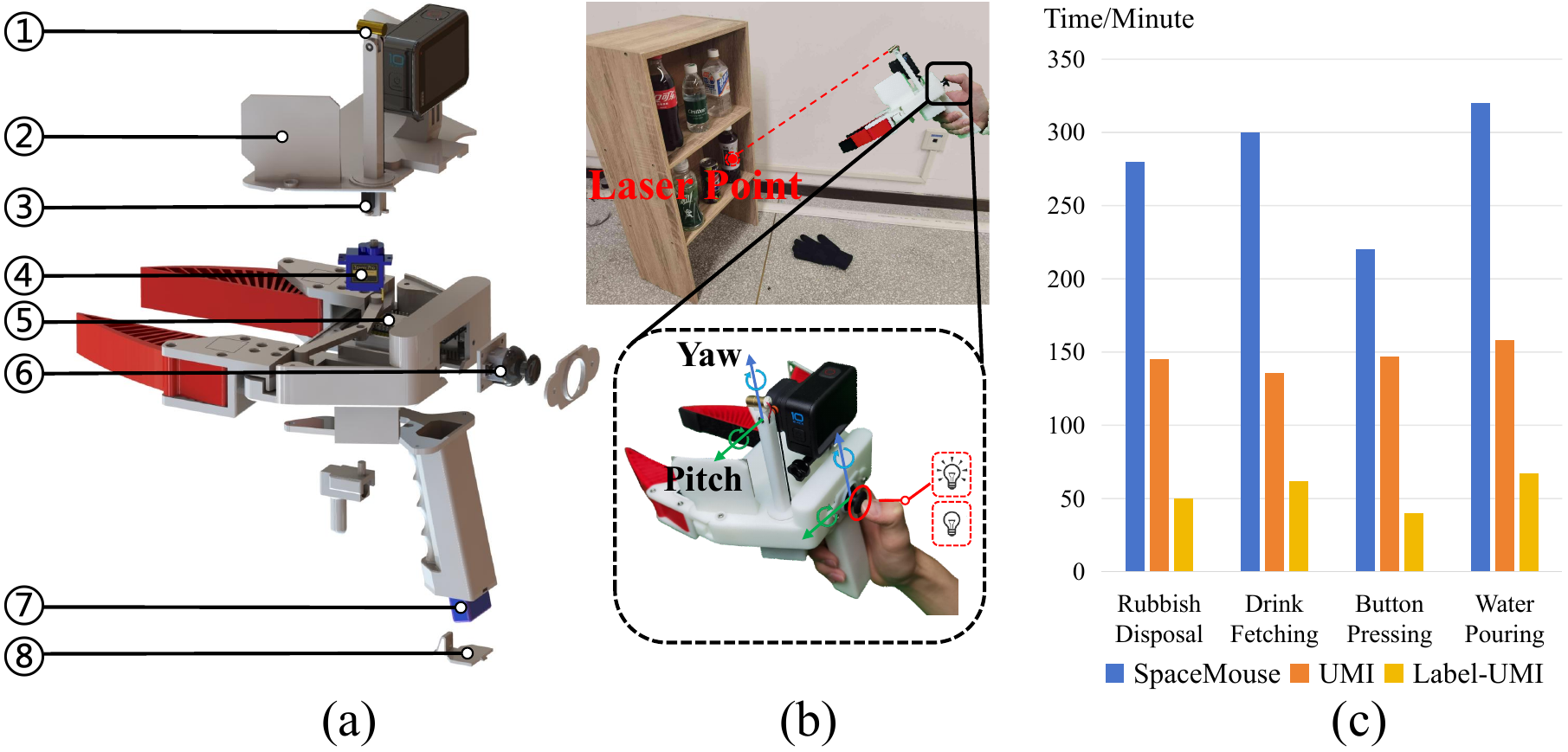}
    \caption{(a) Structure of Label-UMI: \ding{192}Laser, \ding{193}Mirror, \ding{194}Mini servo motor, \ding{195}SG90 servo motor, \ding{196}ESP32 microcontroller, \ding{197}PS2 joystick, \ding{198}Battery, \ding{199}U-shaped bayonet mount. (b) Data collection procedure. (c) Time to collect and annotate 100 samples across different devices.
}
    \label{Structure of Label-UMI}
    \vspace{-0.8CM}
\end{figure}

\vspace{-0.2cm}
\subsection{Integrated Data Acquisition and Processing Pipeline}
Our framework requires efficiently building a dataset in which target objects are annotated with bounding boxes for object localization and grasp planning. However, existing manipulation datasets lack enough environments and objects per task to meet this need. Since UMI \cite{chi2024universal} has proven effective for collecting data to train diffusion policies, we design Label-UMI, an ergonomic device inspired by UMI.

This design ensures accurate baseline reproduction and comparability. As illustrated in Fig. \ref{Structure of Label-UMI}(a), the Label-UMI comprises a compact gimbal with a laser emitter and a PS2 joystick controller; the laser provides precise target positioning and segmentation point cues. Users control the laser's pitch and yaw via the joystick and toggle it on/off with the joystick button (Fig. \ref{Structure of Label-UMI}(b)), operating it intuitively---thumb on the joystick, index finger triggering acquisition---like a game controller.

We replace UMI's rack-and-pinion mechanism with a multi-link system, reducing weight and freeing internal space for microcontrollers, servos, and other electronics, and integrate the power unit into the handle with a U-shaped bayonet mount for easy battery replacement, improving portability and field usability.
To collect data with segmentation markers, the user aligns the laser with the target, starts recording, states the object name into the GoPro microphone, and executes the task, ensuring synchronized data and annotation capture.

The data processing pipeline (Fig. \ref{Overview of Data Acquisiton pipeline.}) automates the extraction of bounding box labels and trajectory data. After collection, audio is transcribed to obtain object labels, while the first video frame is processed by LaserPoint-YOLOv8s—a YOLOv8s model fine-tuned for laser point detection—to locate the laser dot. Its center coordinates serve as a point prompt input along with the video into Segment Anything Model 2 (SAM2), which generates per-frame object masks. Minimum bounding boxes are derived from these masks.
This automated pipeline yields high-precision bounding box annotations for each object in every frame, used to train a real-time YOLO detection model. The approach offers three key advantages:


\begin{itemize}
    \item \textbf{High labeling efficiency}: The entire labeling process is fully automated via script files, thereby eliminating the need for manual label assignment(as shown in Fig.\ref{Structure of Label-UMI}(c)).
    \item \textbf{Fast inference speed}: Compared with zero-shot object detection models like Ground DINO, our model achieves faster inference due to its fewer parameters, making it ideal for real-time robotic tasks.
    \item \textbf{High labeling accuracy}: Our pipeline utilizes SAM2's high accuracy and robustness to achieve fast and precise object segmentation in complex scenes, showcasing strong performance and wide applicability.
    
\end{itemize}

Motion trajectories are processed using ORB-SLAM3, following the UMI framework \cite{chi2024universal}, and are not detailed here. After dataset construction, we employ it to train two key components: a \emph{Bounding-Box Detection Module} and \emph{a Bounding-Box Guided Diffusion Policy}.

\begin{figure*}
    \vspace{0.2cm}
    \centering
    \includegraphics[width=0.85\linewidth]{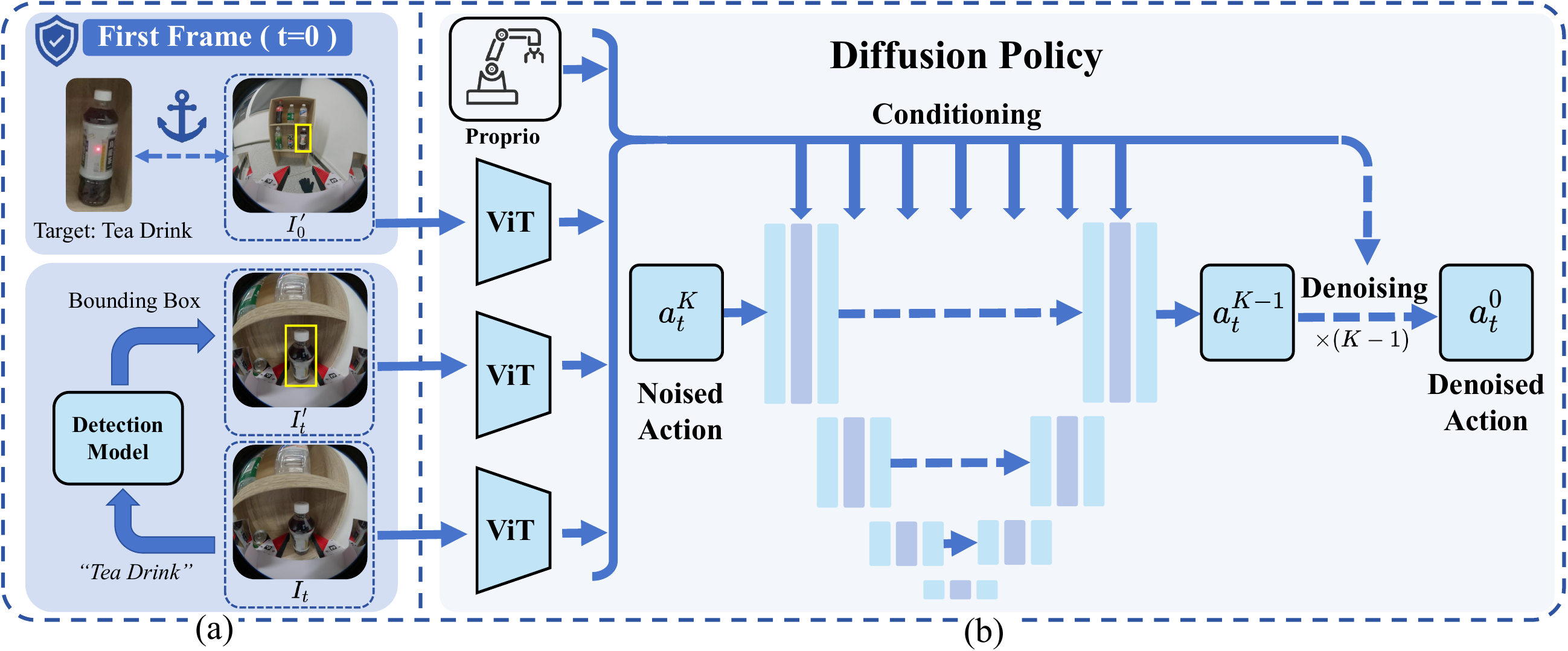}
    \caption{Overview of the BBox-DP. (a) Semantic detection: the raw image and the YOLO-annotated image are encoded separately by a ViT, while the annotated first frame $I'_0$ is fed as a constant input throughout the episode to anchor the target object, via the \emph{First-Frame Anchoring} mechanism. (b) Main policy: the visual features are fused with the robot's proprioceptive state into a unified condition that guides a U-Net diffusion model to produce the action.
}
    \label{Overview of the BBox-DP}
    \vspace{-0.65cm}
\end{figure*}

\subsection{Bounding-Box Detection Module} 

Detection models such as DINOv2 provide accurate, robust bounding boxes, but their high computational cost limits real-time robotic deployment. Such zero-shot models also struggle with abstract or out-of-distribution objects and show limited discrimination for fine-grained categories---e.g., distinguishing beverage brands---despite reliably recognizing general categories like “drink.”

To balance accuracy and efficiency, we adopt a YOLO-based detector trained with data generated by our automated pipeline. This approach achieves real-time performance with high accuracy while maintaining flexibility. By decoupling semantic detection from the main policy into a modular visual instruction interface, the detection module can be seamlessly replaced in the future with more advanced models that offer improved speed or precision, without affecting the overall inference performance of the system.

We distinguish two stages. During training, bounding boxes are generated by our \emph{Data Acquisition and Processing Pipeline} (Fig.\ref{Overview of Data Acquisiton pipeline.}), which uses a laser to localize and annotate the objects. 
During inference, we deploy a YOLO model trained via the same pipeline; owing to our decoupled architecture, the target's box may instead come from other modalities---eye-gaze, an LLM interpreting language, or external detectors---as long as detection meets task requirements. The policy is moreover not tied to per-frame detection: the First-Frame Anchoring mechanism (Section~\ref{Learning Algorithm: BCDP}) keeps it bound to the target even under intermittent detector failures.



\subsection{Bounding-Box Guided Diffusion Policy} 
\label{Learning Algorithm: BCDP}



After training the detection module, we now turn to the core of our framework—the policy model. We introduce an end-to-end imitation learning framework, termed \textbf{BBox-DP}, which extends diffusion policies to model multi-modal action distributions conditioned on visual observations enriched with bounding box annotations, as illustrated in Fig.\ref{Overview of the BBox-DP}.

Specifically, at time step $t$, the semantic instruction $L$ is first processed by an object detection model, which highlights the target objects by generating one or more bounding boxes on the input image $I_t$.
For tasks involving multiple objects (e.g., rearrangement or tool use), different bounding box colors can be used to distinguish between objects with different semantic roles or interaction requirements.
This process produces an augmented image $I'_t$ in which every relevant object is explicitly highlighted.
Both the original image $I_t$ and the annotated image $I'_t$ are subsequently fed into a CLIP-pretrained Vision Transformer (ViT-B/16) for feature extraction. Here, $I_t$ represents the raw RGB image of the physical environment, which contains object contour information essential for determining the grasping posture, whereas $I'_t$ explicitly represents the relative spatial locations of the target objects.

To improve robustness against intermittent detection failures, we introduce a \emph{First-Frame Anchoring} mechanism. At the start of each rollout, we retain the annotated first frame $I'_0$---the initial observation overlaid with the target's bounding box---and feed it as a constant conditioning input throughout the episode. This persistent anchor binds the policy to the target from the outset: even when the detector misses the target in intermediate frames, the policy relies on $I'_0$ to preserve a consistent notion of the target's identity and location, continuing execution instead of stalling on a missing box. It thus downgrades the hard dependence on per-frame detection to a soft one. To further reinforce this, during training we randomly drop or mislabel the boxes in $I'_t$, forcing the policy to tolerate missing or erroneous detections rather than overfitting to perfect annotations.


The extracted features explicitly supply positional cues of the target object, which directly guide the diffusion policy to generate action trajectories toward the object over $K$ denoising steps. We utilize a CNN-based U-Net $\varepsilon_{\theta}$ as the noise prediction network and adopt DDIM \cite{song2020denoising} to reduce inference latency, thereby enabling real-time control. The policy is trained using the loss function defined in Eq. (\ref{loss of diffusion}). This design not only improves the interpretability of the visual representations but also improves the generalization of the model to novel, previously unseen objects.
\begin{align}
    \label{loss of diffusion}
    \mathcal{L}=\operatorname{MSE}\left(\varepsilon^{k}, \varepsilon_{\theta}\left(\mathbf{O}_{t}, \mathbf{a}_{t}^{0}+\varepsilon^{k}, k\right)\right)
\end{align}
where $\varepsilon^k$ is the noise added at diffusion step $k$, $\mathbf{O}_t$ includes the raw image $I_t$, the bounding-box annotated image $I'_t$, the annotated first-frame anchor $I'_0$, and proprioceptive state information $\textit{proprio}$, and $k$ denotes the current denoising iteration step.

\begin{figure*}
    \vspace{0.2cm}
    \centering
    \includegraphics[width=\linewidth]{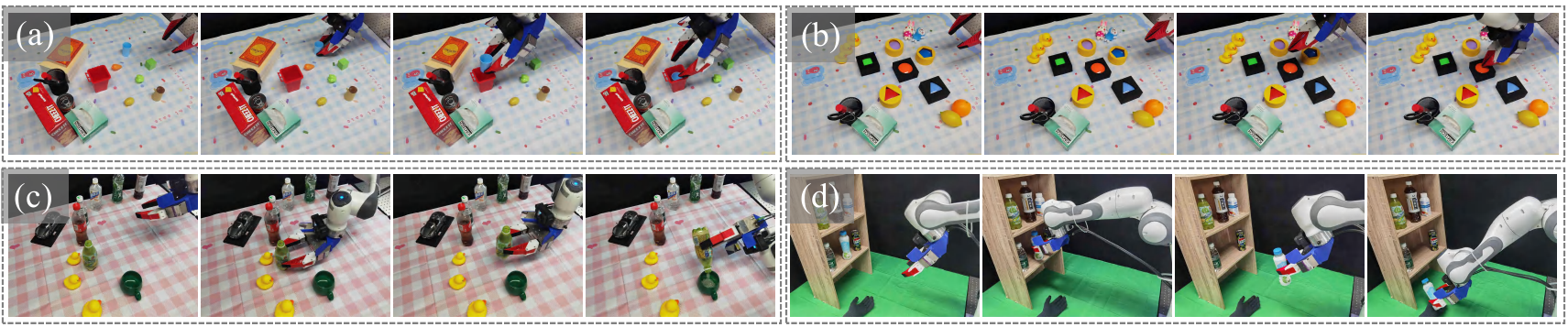}
    \caption{Real-robot experiments on four semantic manipulation tasks: (a) \textbf{Rubbish Disposal}, discarding a specified item into a bin; (b) \textbf{Button Pressing}, pressing a designated button among similar distractors; (c) \textbf{Water Pouring}, pouring a target container's contents into a cup; and (d) \textbf{Drink Fetching}, retrieving a prompted drink from a shelf and handing it over.}
    \label{Real Robot Experiment}
    \vspace{-0.4cm}
\end{figure*}

\section{FORMALIZATION OF DATA SCALLING WITH BOUNDING BOXES}

Building on prior work in data scaling for robotic imitation learning \cite{lin2024data}---where generalization $S$ is modeled as a function of environments ($M$), object instances ($N$), and demonstrations per object–environment pair ($K$), with each object $O_i$ in environment $E_j$ paired with $K$ demonstrations ($D_{ij}^1,\dots,D_{ij}^K$) amid arbitrary distractors---we extend the framework to study how increasing the number of bounding-box objects improves generalization across diverse object categories.

Unlike earlier settings restricted to $N$ objects from a single category, we introduce $N'$ arbitrary object classes ($H_1, H_2, \dots, H_{N'}$) without constraining distractors, and annotate each demonstration with bounding boxes for these objects. The policy is then evaluated on unseen environments and objects using the score $S$ (described later). 
Our study is twofold: (1) characterizing how $S$ scales with the number of bounding-box objects $N'$ and demonstrations per class $K$; and (2) identifying efficient data collection strategies for strong generalization.

To ensure reliable evaluation, we adopt three measures. First, policies are tested only on unseen environments and unseen objects. Second, we use human-assigned stage-wise scores (typically 2–3 stages per task; see Section \ref{EXPERIMENTS}), yielding a normalized score that captures nuanced behavior beyond binary success/failure. Third, to minimize evaluator bias, rollouts from different policies (trained on varying dataset sizes) are randomly interleaved under identical initial conditions, so the evaluator scores each blindly.

\section{EXPERIMENTS}
\label{EXPERIMENTS}

In this section, we first compare our bounding-box-guided policy against other semantics-guided baselines (Sec.~\ref{Comparative Analysis against Semantics-Guided Baselines}). We then stress-test its robustness under missed detections and noisy boxes (Sec.~\ref{Robustness to Detection Failures and Noisy Boxes}). Next, we study how the diversity of bounding-box-annotated objects affects generalization, revealing a data scaling trend (Sec.~\ref{Scaling Laws for Bounding Box Generalization}). Guided by this trend, we propose an efficient data-collection strategy for generalizable, semantics-centered policies (Sec.~\ref{Efficient Data Collection for Generalizable Policies}), and finally validate its cross-task data efficiency on additional tasks (Sec.~\ref{Cross-Task Validation of Data Efficiency}).

\subsection{Overview of Experiments}
We design four real-world tasks (Fig.\ref{Real Robot Experiment}): Rubbish Disposal, Drink Fetching, Button Pressing, and Water Pouring. All run in cluttered scenes with multiple homogeneous distractors, requiring the robot to identify and manipulate the prompted target.
\noindent For dataset construction, each task uses $M=4$ environments and $N'=16$ object classes, with 25 demonstrations per object in each environment (100 per object), totaling 1600 valid demonstrations per task.

\noindent\textbf{Implementation details.} For fair and reproducible comparison, our policy adopts the same backbone and action space as UMI~\cite{chi2024universal}. Each RGB observation (the raw image $I_t$ and the bounding-box-annotated image $I'_t$) is resized to $224\times224$ and encoded by a CLIP-pretrained ViT-B/16 backbone (patch size $16$, $768$-dimensional features) that is fine-tuned end-to-end, using an observation horizon of two steps. 
The policy predicts an action horizon of $16$ steps, using a CNN-based U-Net (channels $256/512/1024$) as the denoising network with DDIM at inference for real-time control. All diffusion-policy methods (Text-DP, Keypoints-DP, and our BBox-DP) share this backbone, action space, training schedule, and data, differing only in the semantic-conditioning interface, so performance gaps reflect the conditioning format rather than implementation.

\begin{figure*}
    \vspace{0.2cm}
    \centering
    \includegraphics[width=0.8\linewidth]{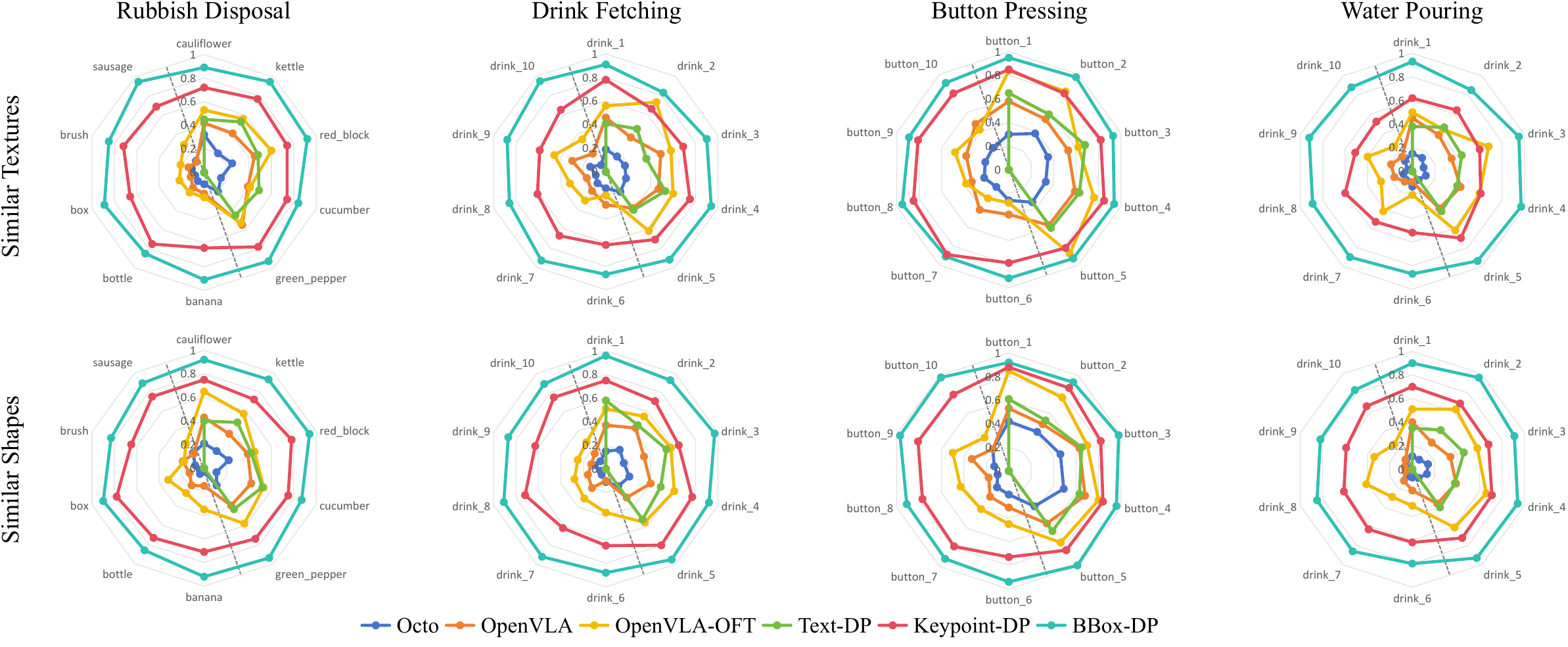}
    \caption{Per-object performance scores. In each radar chart, objects left of the dashed line are the test set and those to the right the training set. Beverage brands and custom buttons, hard to distinguish textually, are shown as codes; the full mapping is in the supplementary video.}
    \label{fig:detail score}
    \vspace{-0.45cm}
\end{figure*}

\begin{table*}[ht]
\caption{Quantitative results across four real-world tasks (Octo, OpenVLA, OpenVLA-OFT, Text-DP, Keypoint-DP, and our BBox-DP). Each model is trained with 4 random seeds; we report mean $\pm$ std of the performance score. Sim. Tex./Sim. Shape denote distractors with similar textures/shapes.}
\vspace{-0.25cm}
\label{Effectiveness study of B-BOXDP in real environments}
\begin{center}
    
\begin{tabular}{@{}c|cccccccc@{}}
\toprule
\multirow{2}{*}{Policy} & \multicolumn{2}{c}{Rubbish Disposal} & \multicolumn{2}{c}{Drink Fetching} & \multicolumn{2}{c}{Button Pressing} & \multicolumn{2}{c}{Water Pouring} \\
                        & Sim. Tex.         & Sim. Shape        & Sim. Tex.        & Sim. Shape       & Sim. Tex.         & Sim. Shape       & Sim. Tex.        & Sim. Shape      \\ \midrule
Octo                    & $16{\pm}4.2\%$   & $13{\pm}3.3\%$   & $14{\pm}4.5\%$  & $13{\pm}3.2\%$  & $28{\pm}5.7\%$   & $30{\pm}5.5\%$  & $11{\pm}3.8\%$  & $9{\pm}3.2\%$ \\
OpenVLA                 & $29{\pm}3.7\%$   & $28{\pm}5.9\%$   & $33{\pm}4.2\%$  & $25{\pm}4.3\%$  & $48{\pm}5.8\%$   & $43{\pm}4.2\%$  & $27{\pm}4.9\%$  & $23{\pm}3.4\%$ \\
OpenVLA-OFT             & $37{\pm}4.3\%$   & $42{\pm}4.7\%$   & $46{\pm}3.1\%$  & $43{\pm}5.7\%$  & $58{\pm}4.6\%$   & $60{\pm}4.4\%$  & $44{\pm}4.6\%$  & $46{\pm}3.6\%$ \\
Text-DP                 & $48{\pm}3.6\%$   & $45{\pm}4.3\%$   & $43{\pm}5.3\%$  & $52{\pm}4.6\%$  & $63{\pm}3.7\%$   & $61{\pm}3.9\%$  & $42{\pm}4.5\%$  & $40{\pm}4.3\%$ \\
Keypoint-DP             & $72{\pm}3.4\%$   & $75{\pm}3.8\%$   & $66{\pm}4.6\%$  & $70{\pm}3.5\%$  & $85{\pm}2.7\%$   & $81{\pm}3.8\%$  & $57{\pm}4.7\%$  & $66{\pm}3.7\%$ \\
\textbf{BBox-DP(ours)}  & $\mathbf{90{\pm}2.1\%}$ & $\mathbf{91{\pm}2.6\%}$ & $\mathbf{89{\pm}3.9\%}$ & $\mathbf{92{\pm}2.8\%}$ & $\mathbf{93{\pm}2.5\%}$ & $\mathbf{95{\pm}2.6\%}$ & $\mathbf{91{\pm}2.8\%}$ & $\mathbf{88{\pm}3.4\%}$ \\ \bottomrule
\end{tabular}

\end{center}
\vspace{-0.8cm}
\end{table*}

For fine-grained evaluation, each task is split into stages. Rubbish Disposal, Drink Fetching, and Water Pouring use three stages---approaching the correct target, grasping it, and completing the final action (discarding, handing over, or pouring)---while Button Pressing uses two (approaching and pressing the designated button). Overall performance is quantified by Eq.~(\ref{the score of traj}):
\begin{equation}
\label{the score of traj}
S = \eta \cdot \frac{1}{n}\sum_{m=1}^{n} S_m
    + \lambda \cdot \text{clip}\!\left(1-\frac{t-t_{\min}}{t_{\max}-t_{\min}},\;0,\;1\right)
\end{equation}
\noindent where the first term is the average stage score where \(S_m \in \{0,1\}\) indicates success at stage \(m\) over \(n\) stages; and the second term measures time efficiency based on the actual completion time \(t\) within thresholds \(t_{\min}\) and \(t_{\max}\); \(\eta,\lambda\) are weighting coefficients with \(\eta+\lambda=1\).


\subsection{Comparison with Semantics-Guided Baselines}
\label{Comparative Analysis against Semantics-Guided Baselines}

To evaluate the necessity of visual object guidance, we compare policy architectures with different semantic-conditioning forms, and add a point-based guidance alternative to assess visual-representation choices. The baselines are:

\begin{itemize}
    \item \textbf{Octo:} A transformer-based robot foundation model \cite{team2024octo}.
    \item \textbf{OpenVLA:} A 7B VLA \cite{kim2025openvla} (Llama-2 + CLIP), a strong full-fine-tuning baseline.
    \item \textbf{OpenVLA-OFT:} A parameter-efficient OpenVLA variant \cite{kim2025fine} using Orthogonal Fine-Tuning.
    \item \textbf{Text-DP:} A text-conditioned diffusion policy \cite{chi2023diffusion} that fuses text and image via FiLM, without bounding-box guidance.
    \item \textbf{Keypoints-DP:} Our variant using 2D keypoints (SAM2 mask centroids) as visual guidance \cite{stone2023open}.
\end{itemize}


For each baseline, we sampled five operational objects from the training set and five from the test set, and tested each policy over 30 randomized trials per object, scoring the results with Eq.~(\ref{the score of traj}). Text-DP was evaluated only on the five training-set objects, as it lacks internet-scale pretraining. All pretrained baselines (Octo, OpenVLA, OpenVLA-OFT) were fine-tuned on our task dataset, and every method used an identical protocol: the same objects, initial conditions, and scoring. We further used two distractor types: objects with similar shapes but different textures, and objects with similar textures but different shapes.

Results are summarized in Table~\ref{Effectiveness study of B-BOXDP in real environments}, with per-object scores in Fig.~\ref{fig:detail score}. Our BBox-DP consistently outperforms all baselines. Specifically, it surpasses Keypoint-DP (point-based guidance), which in turn surpasses Text-DP (no visual guidance): visual guidance provides explicit spatial cues that text cannot, especially in clutter. Beyond point localization, bounding boxes also encode object extent, giving cues for gripper width and reducing grasp failures, making them a richer and more robust visual instruction. A Welch's t-test yields p-values of $3.363\times10^{-7}$ (Rubbish Disposal), $9.736\times10^{-4}$ (Drink Fetching), $4.057\times10^{-4}$ (Button Pressing), and $7.523\times10^{-3}$ (Water Pouring), rejecting the null hypothesis that box guidance does not improve performance. BBox-DP also outperforms the pretrained open-source models, confirming the effectiveness of bounding boxes as “visual instructions.”

\vspace{-0.15cm}
\subsection{Robustness to Detection Failures and Noisy Boxes}
\label{Robustness to Detection Failures and Noisy Boxes}

Any detection-conditioned policy risks stalling when the upstream detector misses the target or returns inaccurate boxes. To quantify the robustness from First-Frame Anchoring (Section~\ref{Learning Algorithm: BCDP}), we stress-test the trained policy by artificially corrupting the detection stream at inference, keeping detector and policy weights unchanged.

\begin{table}[t]
\vspace{0.2cm}
\caption{Robustness to detection corruption, reported as the average performance score (\%) over Rubbish Disposal and Drink Fetching. \emph{w/o FFA} removes the First-Frame Anchoring mechanism, whereas \emph{Ours (Full)} enables it. Best per row in bold.}
\label{tab:robustness}
\centering
\small
\begin{tabular}{@{}l|c|cc@{}}
\toprule
Corruption & Level & w/o FFA & Ours (Full) \\ \midrule
Clean (reference) & --   & 89.4 & 90.5 \\ \midrule
\multirow{3}{*}{\begin{tabular}[c]{@{}l@{}}Missed\\ Detection\end{tabular}}
                  & 20\% & 71.2 & \textbf{89.6} \\
                  & 40\% & 48.3 & \textbf{86.9} \\
                  & 60\% & 25.7 & \textbf{80.8} \\ \midrule
\multirow{3}{*}{\begin{tabular}[c]{@{}l@{}}Noisy\\ Box\end{tabular}}
                  & 0.1  & 80.4 & \textbf{90.1} \\
                  & 0.2  & 60.1 & \textbf{88.5} \\
                  & 0.3  & 38.6 & \textbf{84.4} \\ \bottomrule
\end{tabular}
\vspace{-0.6cm}
\end{table}

We consider two corruption types that emulate realistic detector failures. (i) \emph{Missed detections}: the per-frame bounding box in $I'_t$ is dropped with probability $p_{\text{drop}}\in\{20\%,40\%,60\%\}$, leaving the corresponding frame unannotated, so that a higher $p_{\text{drop}}$ simulates a less reliable detector. (ii) \emph{Noisy boxes}: each box is perturbed by a random translation and rescaling whose magnitude is a fraction $\sigma\in\{0.1,0.2,0.3\}$ of the box size, emulating jittery or imprecise localization. We compare the full model against an ablation that removes the First-Frame Anchoring mechanism (\emph{w/o FFA}). All variants are evaluated on the Rubbish Disposal and Drink Fetching tasks under the same protocol as Section~\ref{Comparative Analysis against Semantics-Guided Baselines}, with the clean setting ($p_{\text{drop}}=0$, $\sigma=0$) serving as the reference.

As reported in Table~\ref{tab:robustness}, the ablated model is highly sensitive to detection corruption: its performance score collapses from $89.4\%$ to $25.7\%$ once $60\%$ of the boxes are dropped, and degrades comparably under box jitter, confirming that a naively detection-conditioned policy inherits the brittleness of the detector. In contrast, our full model degrades gracefully, retaining a score of $80.8\%$ even under a $60\%$ drop rate and $84.4\%$ under the strongest jitter ($\sigma=0.3$). On the clean setting, the two variants are statistically indistinguishable, showing that the added robustness comes at no cost to nominal performance. These results confirm that First-Frame Anchoring relaxes the policy's hard dependence on per-frame detection to a soft one, making the system substantially more tolerant of missed and noisy detections.

\subsection{Scaling Laws for Bounding Box Generalization}
\label{Scaling Laws for Bounding Box Generalization}
We study two manipulation tasks, Rubbish Disposal and Drink Fetching, using the same dataset as Section~\ref{Comparative Analysis against Semantics-Guided Baselines} (1600 demonstrations per task). To disentangle the effects of object diversity and demonstration count, we index each configuration by a triple $(m,n)_{j}$: $m\in\{0,\dots,4\}$ sets the number of bounding-box objects to $2^m$ (sampled from a pool of 16), $n\in\{0,\dots,-5\}$ sets the per-object demonstration fraction to $2^n$, and $j$ is the repetition index. We use five random samplings for $m<4$ ($j\in[1,5]$) and a single one for $m=4$ ($j=1$).
Training every valid configuration with more than 50 total demonstrations yields 76 policies. Each is evaluated on 16 unseen objects in an unseen environment over 30 trials per object (480 trials), and its score is averaged over all repetitions $j$. For $m<4$, the five samplings act as different seeds and thus capture the variance from object selection; for $m=4$ (all 16 objects), we instead train five models with different random seeds and report their variance.

\begin{figure}[t]
    \vspace{0.25cm}
    \centering
    \includegraphics[width=\linewidth]{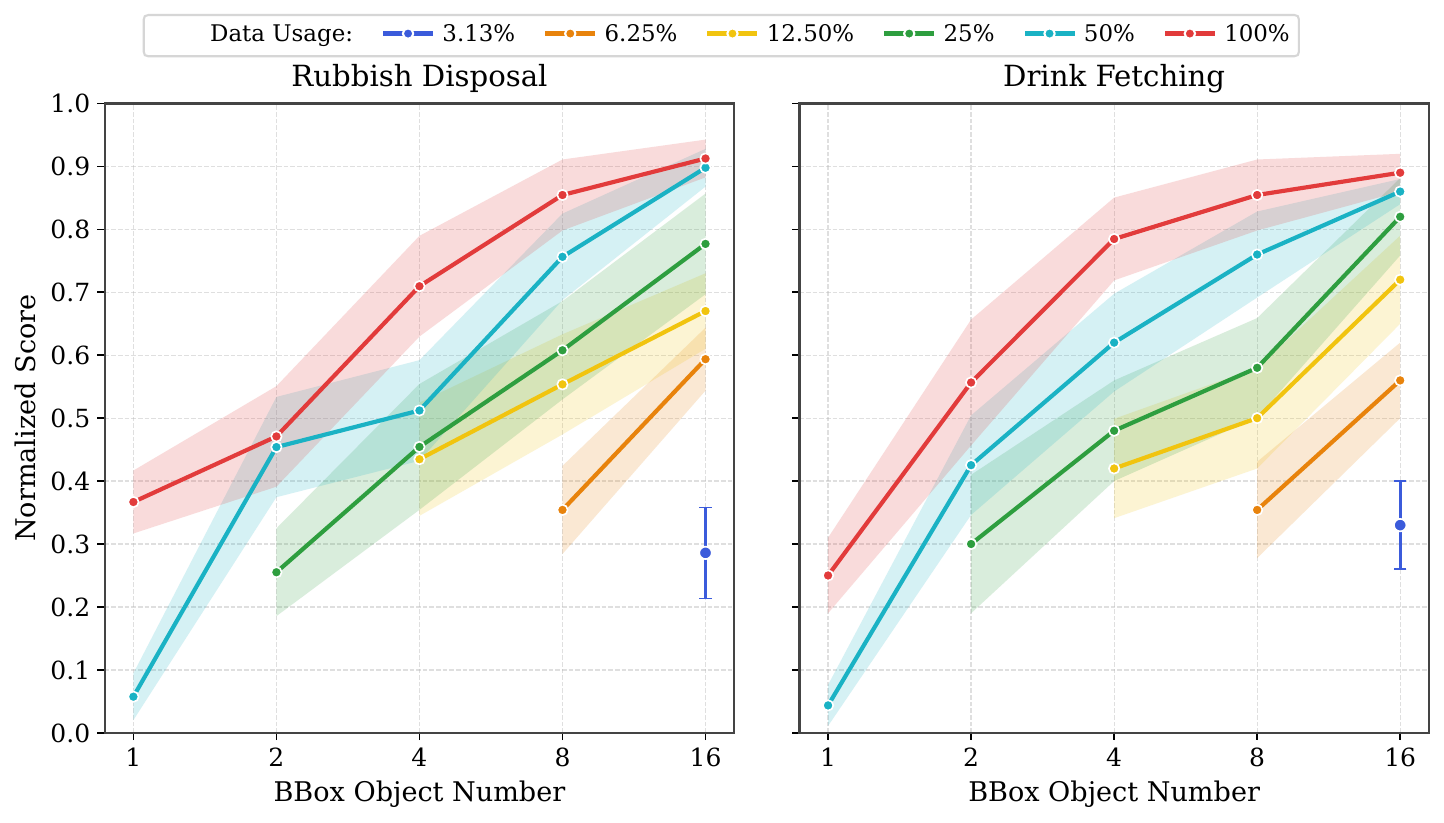}
    \caption{Generalization across Bounding-Box Object Number.}
    \label{fig: Generalization Result.}
    \vspace{-0.7cm}
\end{figure}

As shown in Fig.\ref{fig: Generalization Result.}, we summarize the experimental results for both tasks, revealing two key findings: 
(1) The policy’s generalization performance consistently improves as the number of bounding-box objects increases, across all demonstration fractions. (2) Increasing the number of bounding-box objects reduces the number of demonstrations needed per object. 
For example, in Rubbish Disposal, using 4 bounding-box objects produces a noticeable performance gap between 50\% and 100\% demonstration fractions, whereas this gap nearly disappears when 16 objects are used.

\begin{figure}[t]
    \vspace{0.2cm}
    \centering
    \includegraphics[width=\linewidth]{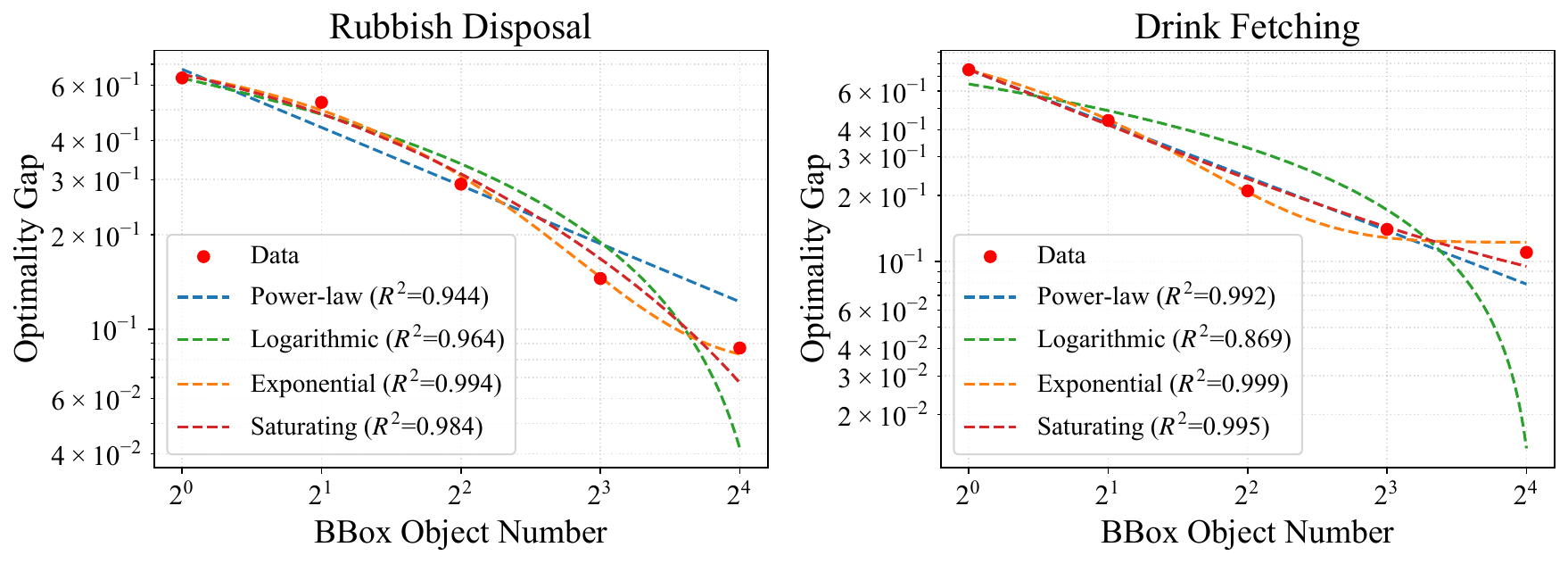}
    \caption{Fitted scaling models of the optimality gap versus the number of bounding-box objects on the two tasks. Quantitative comparison is reported in Tab.~\ref{tab:scaling-fit}.}
    \label{fig:Power-law}
    \vspace{-0.5cm}
\end{figure}

Then we examine how the optimality gap $Y=1-\text{Normalized Score}$ scales with the number of bounding-box objects $X$. Rather than assuming a single functional form, we fit four candidate models in the original (non-logarithmic) space---a power law $Y=\beta X^{\alpha}$, a logarithmic, an exponential, and a saturating model with a positive lower bound---and compare them by $R^2$, AIC, and BIC.
As shown in Tab.~\ref{tab:scaling-fit} and Fig.~\ref{fig:Power-law}, the exponential and saturating models fit best on both tasks. Generalization thus improves monotonically with object diversity but with \emph{diminishing marginal gains}, saturating toward its upper bound (the normalized score cannot exceed $100\%$) rather than growing unbounded as a power law implies. A log-log power-law fit still shows a strong linear correlation ($r\approx-0.99$; exponents $\alpha=-0.76\,[-0.99,-0.53]$ and $-0.72\,[-0.96,-0.48]$ at the $95\%$ confidence level), but the curvature in Fig.~\ref{fig:Power-law} reveals that this description holds only approximately over the tested range of one to sixteen objects.
Crucially, all four models agree on this monotonic-yet-diminishing trend, which underlies our data collection strategy. Unlike the unbounded two-parameter power law of Lin et al.~\cite{lin2024data}, which by their own account becomes unreliable beyond the fitted range, our bounded formulation better reflects this saturating behavior. 


\subsection{Efficient Data Collection for Generalizable Policies}
\label{Efficient Data Collection for Generalizable Policies}

\begin{table}[t]
\vspace{0.2cm}
\caption{Model comparison for the data scaling trend (lower AIC is better; best per task in bold).}
\label{tab:scaling-fit}
\centering
\small
\begin{tabular}{@{}l|cc|cc@{}}
\toprule
 & \multicolumn{2}{c|}{Rubbish Disposal} & \multicolumn{2}{c}{Drink Fetching} \\
Model & $R^2$ & AIC & $R^2$ & AIC \\ \midrule
Power-law   & 0.944 & $-23.9$          & 0.992 & $-32.6$          \\
Logarithmic & 0.964 & $-26.1$          & 0.869 & $-18.4$          \\
Exponential & \textbf{0.994} & $\mathbf{-32.7}$ & \textbf{0.999} & $\mathbf{-40.3}$ \\
Saturating  & 0.984 & $-28.2$          & 0.995 & $-32.7$          \\ \bottomrule
\end{tabular}
\vspace{-0.6cm}
\end{table}

We now address a practical question: \emph{given a task, how should one choose the number of annotated object classes ($N'$) and demonstrations per object ($K$) to maximize generalization at minimal collection effort?}

\noindent\textbf{How much to collect.} As shown in Section~\ref{Scaling Laws for Bounding Box Generalization}, once the performance score exceeds 50, adding object diversity yields larger gains than adding demonstrations---most clearly at $K=50$, where performance rises fastest and reaches about $85\%$ at 16 objects. We therefore recommend prioritizing a diverse object set (around 16 classes) with roughly 50 demonstrations each. This balances performance against collection cost; only when diversity is capped by resource limits and peak performance is essential should one add more demonstrations per object.

\noindent\textbf{Which objects to collect.} With $m$ and $n$ fixed, performance still varies considerably across object sets, correlating strongly with the shape variation within a set---especially during the final approach, when the arm nears the target to grasp. To probe this, we set $m=3,\,n=0$ (8 objects, full demonstrations) and generated 20 random subsets, then trained and evaluated 20 corresponding policies. Generalization depends on both the uniformity of object counts across shape groups and the proportion of distinct shape groups. Leaving the exact mechanism to future work, we distill a practical guideline: \emph{for tasks with large shape variation (e.g., Rubbish Disposal), training on highly shape-diverse objects markedly improves performance, whereas for tasks with limited shape variation (e.g., Drink Fetching), performance is far less sensitive to shape diversity.}

\begin{table}[t]
\vspace{0.2cm}
\caption{Success rate across all tasks.}
\label{Success rate across all tasks}
\centering
\begin{tabular}{@{}c|cccc@{}}
\toprule
Task             & \begin{tabular}[c]{@{}c@{}}Rubbish\\ Disposal\end{tabular} & \begin{tabular}[c]{@{}c@{}}Drink\\ Fetching\end{tabular} & \begin{tabular}[c]{@{}c@{}}Button\\ Pressing\end{tabular} & \begin{tabular}[c]{@{}c@{}}Water\\ Pouring\end{tabular} \\ \midrule
Score        & 89.78                                                   & 86.45                                                 & 91.21                                                  & 87.36                                                \\
Success Rate & 90.45\%                                                 & 87.43\%                                               & 92.26\%                                                & 86.79\%                                              \\ \bottomrule
\end{tabular}
\vspace{-0.6cm}
\end{table}

\vspace{-0.15cm}
\subsection{Cross-Task Validation of Data Efficiency}
\label{Cross-Task Validation of Data Efficiency}

To verify the general applicability of our data collection strategy, we applied it to two additional tasks:  Button Pressing and Water Pouring. 
Data collection followed the efficient strategy derived from Section \ref{Efficient Data Collection for Generalizable Policies}. As indicated in Table \ref{Success rate across all tasks}, the trained policies achieve success rates of around 85\% across all four tasks, including both the previously studied tasks and the two newly introduced ones.

\section{CONCLUSIONS}
In this work, we proposed a bounding-box guided policy framework that integrates the Label-UMI data collection device and the BBox-DP policy, which leverages visual object cues (bounding-box) to enhance generalization in semantic manipulation. 
Characterizing how performance scales with object diversity, we revealed a data scaling law with clearly diminishing, saturating returns, which motivates an object-diversity-first strategy for efficient dataset design.
Extensive real-world experiments confirm its effectiveness: our method sustains over 85\% success across diverse tasks, even under challenging texture- and shape-similar distractors, and remains robust to missed and noisy detections.
Overall, this work offers a scalable solution for semantic manipulation and points to promising directions for data-efficient robot learning.
In future work, we will study how object shapes in the dataset affect policy performance and explore more robust visual instructions for cluttered and occluded settings.







\vspace{-0.3cm}
\bibliographystyle{IEEEtran}
\bibliography{IEEEabrv,mylib}

\end{document}